\documentclass[letterpaper, 10 pt, conference]{IEEEtran}

\IEEEoverridecommandlockouts

\usepackage{fullpage, booktabs, multirow, multicol, graphicx, subcaption, url, float, tabularx}
\usepackage{cite}
\usepackage[bookmarks=true]{hyperref}
\usepackage{xcolor}

\usepackage{amsmath,amssymb,amsthm,mathrsfs,bm,pifont}
\usepackage{scalerel,stackengine}
\usepackage{algpseudocode, algorithm}
\theoremstyle{nospace} \newtheorem{remark}{Remark}

\usepackage[subtle, title]{savetrees}
\newcommand{\U}{\mathcal{U}}
\newcommand{\reals}{\mathbb{R}}

\newcommand{\stale}{{\bf SB}}
\newcommand{\mastale}{{\bf MA-SB}}
\newcommand{\cde}{{\bf InFuser}}
\newcommand{\ndp}{{\bf NDP}}
\newcommand{\rmax}{r_{\mathrm{max}}}
\newcommand{\rtot}{r_{\Sigma}}
\newcommand{\ie}{i.e., }
\newcommand{\eg}{e.g., }

\newcommand{\rev}[1]{\textcolor{black}{#1}}

\newtoggle{arxiv}
\toggletrue{arxiv}

\graphicspath{{./fig/}}

\title{\LARGE \bf 
Multiscale Sensor Fusion and \\\vspace{0.2 cm} Continuous Control with Neural CDEs
}
\date{}

\author{Sumeet Singh$^{1}$, Francis McCann Ramirez$^{2}$, Jacob Varley$^{1}$, Andy Zeng$^{1}$, and Vikas Sindhwani$^{1}$
\thanks{$^{1}$Robotics at Google, NYC;
        Corresponding email: {\tt\small ssumeet@google.com}}%
\thanks{$^{2}$Google AI Resident, NYC}%
}

\begin{document}

\maketitle

\begin{abstract}
Though robot learning is often formulated in terms of discrete-time Markov decision processes (MDPs),  physical robots require near-continuous multiscale feedback control. Machines operate on multiple asynchronous sensing modalities, each with different frequencies, \eg video frames at 30Hz, proprioceptive state at 100Hz, force-torque data at 500Hz, etc. While the classic approach is to batch observations into fixed-time windows then pass them through feed-forward encoders (\eg with deep networks), we show that there exists a more elegant approach -- one that treats policy learning as modeling latent state dynamics in continuous-time.
Specifically, we present $\cde$, a unified architecture that trains continuous time-policies with Neural Controlled Differential Equations (CDEs). $\cde$ evolves a single latent state representation over time by (In)tegrating and (Fus)ing multi-sensory observations (arriving at different frequencies), and inferring actions in continuous-time.
This enables policies that can react to multi-frequency multi-sensory feedback for truly end-to-end visuomotor control, without discrete-time assumptions.
Behavior cloning experiments demonstrate that $\cde$ learns robust policies for \emph{dynamic} tasks (e.g., swinging a ball into a cup) notably outperforming several baselines in settings where observations from one sensing modality can arrive at much sparser intervals than others.

\end{abstract}

\section{Introduction}

The task of registering readings from multiple asynchronous sensing modalities (\eg 30Hz video frames, 100Hz proprioceptive state, 500Hz force-torque data), fusing them as soon they arrive to form contact-rich representations of the environment, and integrating them over time, is key to enabling robots to adapt swiftly and precisely to the dynamics of unstructured environments.
This is particularly important for agile behaviors where feedback control needs to be nearly continuous-time.

\begin{figure}[t]
  \centering
  \vspace{0.5em}
  \includegraphics[width=0.99\linewidth]{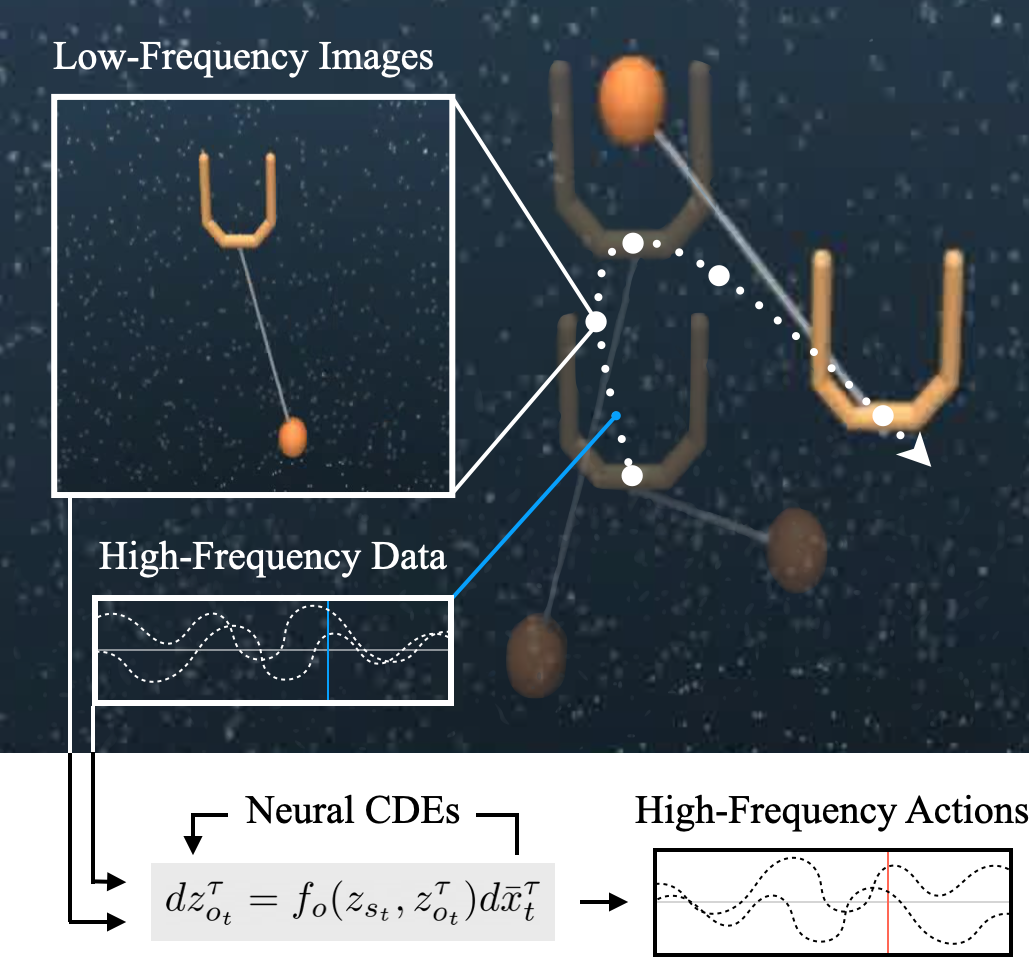}  
\caption{\footnotesize{{\it $\cde$ uses Neural CDEs to learn quasi continuous-time policies that can react to multi-frequency multi-sensory feedback (\eg low-frequency images, high-frequency proprioceptive force-torque data), and integrates temporal information over observations to infer high rate actions.}}}
\label{fig:figpage1}
\vspace{-3mm}
\end{figure}
In this work, we develop a novel policy architecture grounded in the formalism of Neural Controlled Differential Equations (CDEs) \cite{kidger2020neural}. CDEs are a generalization of Neural ODEs \cite{chen2018neural},
whereby one evolves a latent state against a \emph{process} as opposed to only time. As depicted in Figure~\ref{fig:figpage1}, we leverage a Neural CDE model that is \emph{conditioned} on images and \emph{driven} by the set of sensing modalities arriving at a higher frequency than the images. Our method yields policies that can react to multi-frequency multi-sensory feedback for truly end-to-end visuomotor control, without any discrete-time assumptions or having to pre-train a family of (feedback) primitives.


\noindent\textbf{Related Work}: {\it Multi-sensory fusion} with deep networks have shown promising results in learning cross-modal representations for discriminative and generative tasks \cite{lee2019making,li2019connecting,lee2021detect,liu2020understanding,gao2021objectfolder,yang2021learning,calandra2018more,hu2021unit,sung2017deep,de2018integrating}. These architectures typically consist of independent modality-specific encoders, for instance, convolutional neural networks (CNNs) for transient inputs such as images and depth perception, recurrent neural networks (RNNs) for temporally correlated inputs such as IMU sensors and audio inputs, and more recently, transformers for natural language. Embeddings generated from these independent encoders are fused into a single multi-sensory latent representation that is then used for downstream tasks. Training these networks has been addressed via both end-to-end methods (i.e., guided by the downstream task objective) \cite{liu2020understanding,gao2021objectfolder,yang2021learning,calandra2018more,de2018integrating} and pre-training techniques \cite{lee2019making,li2019connecting,hu2021unit,sung2017deep,lesort2018state}, the latter motivated from the perspective of \emph{representation learning}. In the context of robot control however, such architectures implicitly assume a discrete-time MDP, by batching past observations with fixed time windows.

Imbuing learnable policies with {\it dynamical systems-based structures} has also been a critical area of research within robot control, with Dynamic Movement Primitives (DMPs) being one of the most predominant methodologies; see \cite{saveriano2021dynamic} for an extensive review. DMPs fall within the general class of stable dynamical systems-based imitation learning, whereby one fits nonlinear dynamical systems to observed state trajectories provided by an expert, with the constraint that the dynamics satisfy certain stability criteria, e.g., in the sense of Lyapunov~\cite{khansari2011learning, khansari2014learning,khader2021learning}, or contraction~\cite{singh2021learning,sindhwani2018learning,khadir2019teleoperator}. DMPs have been extended to incorporate sensor feedback \cite{rai2017learning,chebotar2014learning}, and used within hierarchical reinforcement learning within an options framework \cite{daniel2012hierarchical, stulp2012reinforcement, parisi2015reinforcement, sutton1999between}. Recently, they have been merged with deep learning whereby the parameters governing the DMPs are generated by an encoder processing higher-dimensional observations (e.g., images) \cite{peters2008reinforcement,bahl2020neural,bahl2021hierarchical}. We note however that while all these methods enable reasoning over the space of trajectories, they have only been demonstrated on datasets with single stream modalities

There has been recent work that jointly addresses multi-sensory fusion and dynamical systems. Motivated from the perspective of decomposing manipulation tasks into a sequence of modality-specific phases, \cite{narita2021policy} learns a family of DMP policies utilizing various subsets of different sensing modalities (RGB, force-torque, proximity images), where the primitive forcing functions are modulated directly by the sensor feedback. An overall ``blending" policy network is trained to smoothly combine all the DMP policies to produce the overall control action. \cite{escontrela2020zero} leverages a multi-sensory encoder (processing exteroceptive and proprioceptive inputs) to update the parameters of a periodic trajectory generator governing legged locomotion.

\subsection*{Statement of Contributions}
Our key observation is that while multi-sensory fusion and \rev{dynamical systems-based policies} have been predominantly studied in isolation, they are two tightly intertwined objectives that can be modeled jointly with a single continuous-time architecture. Leveraging the universality of CDEs in representing functions over irregularly sampled time-series~\cite{kidger2020neural}, we construct a hybrid continuous-time model that \rev{uses multi-frequency observations to drive the evolution of a continuous-time multi-sensory latent embedding. This model is trained using the adjoint-based backpropagation method associated with Neural ODEs to generate continuous-time action \emph{functions}}. While the generality of the model enables training with either reinforcement learning or imitation, in this work we focus on the latter, in particular, behavior cloning. We train the CDE-based model and appropriate baselines on expert data generated from increasingly challenging (simulated) environments, and quantify the learned policies' performance under various deployment settings such as sensor throttling and missed packets. The experiments demonstrate that on quasi-static tasks such as cloth manipulation, the CDE-based architecture is competitive with existing state-of-the-art multi-sensory fusion policies. However, with increasing task dynamism and sparser image rates, the CDE-based architecture outperforms explicit models that do not feature any temporal abstraction as well as recent state-of-the-art deep DMP-based policies~\cite{bahl2020neural}.


\section{Policy Learning}
Policy learning can be formulated in a variety of ways, and typically assumes the discrete-time MDP setting: where observations $o_t \in \mathcal{O}$ at discrete time-index $t \in \mathbb{N}$ are mapped to actions $a_t \in \mathcal{A}$, by a feedback policy $\pi_{\theta}: {\cal O} \to {\cal A}$, where $\theta$ are learnable parameters (\eg of a neural network).
Reinforcement learning (RL) methods seek an optimal closed-loop policy by maximizing the expected sum of discounted rewards, where the robot applies $\pi_{\theta}$ recursively through robot-environment transition dynamics (the distribution of which can also be learned) collecting state-wise rewards.
Meanwhile, imitation learning (IL) via behavioral cloning formulates policy learning as an instance of supervised learning on expert demonstrations. Given observation-action pairs, this entails solving a regularized loss minimization problem of the standard form,
$$\theta^* = \min_{\theta} \sum_{i, t} l(\pi_{\theta} (o^i_t), a^i_t) + \Omega(\theta)$$ where $l(\cdot, \cdot)$ is imitation loss and $\Omega$ is a regularizer.  Behavioral cloning can be effective~\cite{florence2022implicit, zeng2020transporter} when prediction errors do not compound significantly over time or are mitigated using DAGGER-like techniques~\cite{jang2022bc}. 



\subsection{Hybrid Continuous-Time Policies}

In this work, we study the hybrid continuous-time (HCT) setting, structured as follows. Suppose that we observe image(s) $s_t \in \reals^{H\times W\times C}$ at every discrete time index $t \in \mathbb{N}$, and in-between time indices $t$ and $t+1$, we have access to \emph{continuous-time} (this will be straightforwardly relaxed to ``higher-frequency") observations from other sensing modalities, denoted by the \emph{function} $x_t(\cdot): \tau \in [0, T] \mapsto \reals^n$. The constant $T$ is a user-set parameter that models the length of time in-between successive images.

An HCT policy is defined as a \emph{functional} map from the observation tuple $o_t := (s_t, x_t(\cdot))$ to a control \emph{function} $u_t(\cdot): \tau \in [0, T] \mapsto \U$, where $\U$ is the control space. Therefore, in ``MDP-notation", our action $a_t$, mapped from $o_t$, is the control function $u_t(\cdot)$\footnote{For brevity, for any given function $y_t(\cdot)$ with domain $[0, T]$, we use the notation $y_t^\tau$ to denote $y_t(\tau)$.}, drawing natural analogies with hierarchical reinforcement learning.

\begin{remark}
Although we have described the observation structure as a nesting of higher-frequency observations $x_t(\cdot)$ in-between low-frequency observations $\{s_t, s_{t+1}\}$, one should really interpret the two signals as separate asynchronous sensor streams observed at differing frequencies. 
\end{remark}

To ensure tractability of the functional map $\pi_{\theta}: o_t = (s_t, x_t(\cdot)) \mapsto u_t(\cdot)$, we impose two conditions. First, we constrain the functions $u_t(\cdot)$ to be at-least piecewise $\mathcal{C}^1$ over the interval $[0, T]$. Note that this does not prevent non-smooth transitions when a new image is observed at discrete time index $t+1$. This is a justified assumption for robotics applications where the image update frequency is typically at least 2 Hz, and a smooth control command is required at the (higher) control frequency.

Our second assumption states that the control function $u_t(\cdot)$ is causal w.r.t. the observation $o_t= (s_t, x_t(\cdot))$; \rev{a natural assumption given our goal of deploying a real-time controller}. Formally, let $S_t$ represent the random variable for images $s_t$ sampled at discrete time index $t$, and $X_t^\tau, \tau \in [0, T]$ represent the continuous-time stochastic process for state functions $x_t(\cdot)$ sampled in between discrete time indices $t$ and $t+1$. Causality dictates that the stochastic process $U_t^\tau, \tau \in [0, T]$ with sample realization $u_t(\cdot)$ is
\emph{adapted} to the filtration generated by the random variable $S_t$ and the natural filtration~\cite{oksendal2013stochastic} of $X_t^\tau$.

An elegant modeling framework compatible with these assumptions is where $u_t(\cdot)$ is computed as the solution to a neural controlled differential equation, introduced next.

\subsection{Neural Controlled Differential Equations} \label{sec:cde}

For self-containment, we provide a brief introduction to CDEs. For further detail we refer the reader to~\cite{kidger2020neural,lyons2007differential}. Let $y: [0, T] \rightarrow \reals^n$ be a continuous function with bounded variation, and let $f: \reals^d \rightarrow \reals^{d \times n}$ be a matrix-valued continuous function. We can then define a continuous path $z: [0, T] \rightarrow \reals^d$ as the solution to the following integral:
\begin{equation}
    z(\tau') = z(0) + \int_0^{\tau'} f(z(\tau))\, dy^{\tau}, \quad \tau' \in [0, T].
\label{cde_int}
\end{equation}
This is a Riemann-Stieltjes integral; global existence and uniqueness only requires weak regularity conditions on $f$, namely Lipschitz continuity~\cite{lyons2007differential}. In differential notation, the CDE in~\eqref{cde_int} is written as:
\begin{equation}
    dz^\tau = f(z^\tau)\, d y^\tau,
\label{cde_diff}
\end{equation}
where we remark the similarity with stochastic differential equations notation, owing to the natural analogy between the \emph{driving process}, i.e., $y$, and the \emph{driven process}, i.e., $z$.

As our use of CDEs will be in the context of processing incoming sensor streams, i.e., sequences of time-stamped data $\bm{x}_t:= \{(0, x_t(0)), (\tau_i, x_t(\tau_i)), \ldots, (T, x_t(T))\}$, we use the following time-series adaptation of CDEs. Let $\hat{x}_t: [0, T] \rightarrow \reals^n$ be the natural spline interpolant of the values $\{x_t(\tau_i)\}_i$, and define the map $\bar{x}_t: \tau \in [0, T] \mapsto (\hat{x}_t(\tau), \tau) \in \reals^{n+1}$. Let $f_{\theta}: \reals^d \rightarrow \reals^{d \times (n+1)}$ be any parameterized matrix-valued continuous function. Consider now the following CDE:
\begin{equation}
    dz_t^\tau = f_{\theta}(z_t^\tau)\, d\bar{x}_t^\tau,
\label{ncde}
\end{equation}
where the initial condition $z_t^0$ is a separately parameterized explicit function of $x_t^0$. The solution to equation~\eqref{ncde} is termed a Neural CDE model.

\begin{remark}
    As $\bar{x}_t(\cdot)$ is differentiable in our setting (a natural cubic spline), the CDE is converted into the following ODE:
    \[
        \dfrac{d z_t(\tau)}{d \tau} = f_{\theta}(z_t^\tau) \dfrac{d \bar{x}_t(\tau)}{d \tau},
    \]
    and integrated using an off-the-shelf multi-step integrator. Gradient backpropagation may then be performed using the established adjoint sensitivity technique~\cite{chen2018neural}.
\end{remark}

\rev{Despite sharing the same adjoint-based training technique, Neural CDEs are strictly more general than Neural ODEs. To see this more clearly, consider augmenting $z_t^\tau$ to include $\bar{x}_t^\tau$ and setting the relevant sub-matrix of $f_{\theta}$ as the identity. Then, one would recover Neural ODE representations of the form:
\[
    \dfrac{dz_t(\tau)}{d\tau} = f_{\theta}(z_t(\tau), \bar{x}_t(\tau)).
\]
In general, Neural CDEs constitute an elegant modeling framework for defining ``universal" mappings over the space of time-series $\bm{x}_t$~\cite{kidger2020neural}[Theorem B.14], and may be seen as a generalization of the continuous-time limit of RNNs}. In our framework, we will leverage neural CDEs to evolve a multi-sensory latent embedding, driven by the higher-frequency observation stream $x_t$.

\begin{figure*}[t]
    \centering
    \includegraphics[width=0.95\linewidth]{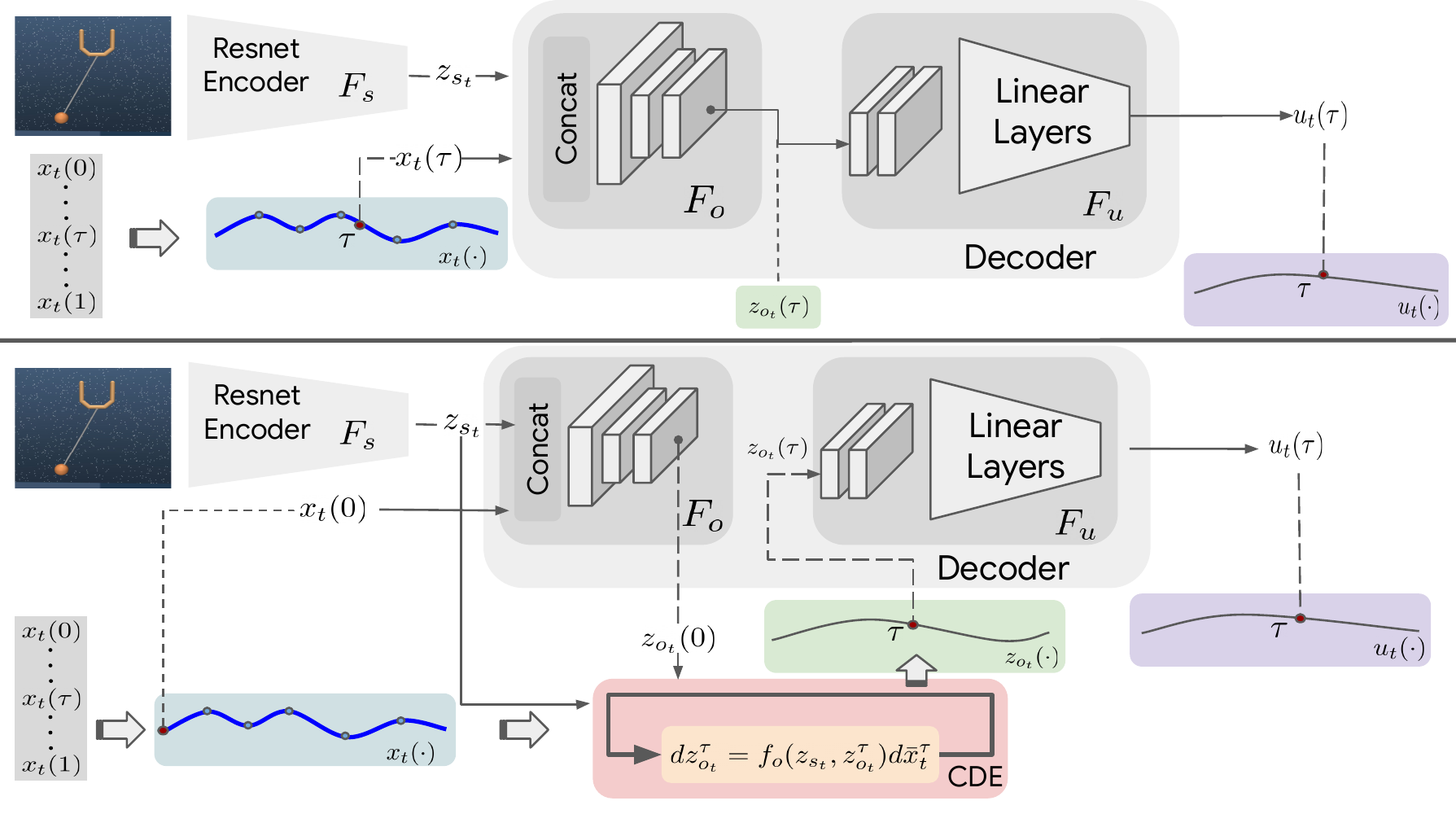}
    \caption{\footnotesize{In the figures above a line with an arrow at the end denotes a single value whereas the striped 2D arrows denote a functional map. \emph{Top}: $\stale$ baseline model; \emph{Bottom}: $\cde$ (proposed). Note that the main difference between the two architectures lies in the middle of the Decoder block. $\cde$ evolves the latent vector $z_{o_t}$ from its initial value $z_{o_t}(0)$, generated in identical fashion to $\stale$, via a neural CDE that is driven by the incoming higher-frequency observations $x_t(\cdot)$. In contrast, the $\stale$ model updates the latent vector $z_{o_t}$ in a stateless manner. Both models decode this latent vector at any intermediate time $\tau$ into the control action using identical MLP structures.}}
    \label{fig:archs}
    \vspace{-3mm}
\end{figure*}

\section{Designing HCT Policies with CDEs}

To introduce our CDE-based model, it is helpful to begin with a baseline architecture that more closely resembles multi-sensory fusion models seen in the literature. We first make precise the nature of the higher-frequency observations $x_t$. In particular, we assume that in between images $s_t$ and $s_{t+1}$, we collect a sequence of time-stamped higher-frequency sensor measurements $\bm{x}_t:= \{(0, x_t(0)), (\tau_i, x_t(\tau_i)), \ldots, (T, x_t(T))\}$, where $\tau = 0$ corresponds to discrete index $t$ and $\tau = T$ corresponds to index $t+1$. 

Consider Figure~\ref{fig:archs} (top), which illustrates our first baseline model, and the following accompanying equations, which we hereby term $\stale$ (stale baseline):
\begin{subequations}
\begin{align}
    z_{s_t} &:= F_s(s_t) \\
    z_{o_t}(\tau) &:= F_o(z_{s_t}, \hat{x}_t(\tau)),\quad \tau \in [0, T) \\
    u_t(\tau) &:= F_u(z_{o_t}(\tau)),\quad \tau \in [0, T)
\end{align}
\label{stale_model}
\end{subequations}
Here, $F_s$ is an image-only encoder mapping the image observation $s_t$ to a latent image embedding $z_{s_t}$, that is subsequently held fixed for all $\tau \in [0, T]$. For each $\tau \in [0, T]$, we feed $z_{s_t}$ and the concurrent (interpolated) observation $\hat{x}_t(\tau)$ into a fusion network $F_o$, to produce a multi-sensor latent embedding $z_{o_t}(\tau)$. This embedding is then decoded via $F_u$ to produce the control value $u_t(\tau)$. Note that such a baseline model encapsulates the most common adaptation of multi-sensory policies for robot control, and serves as a useful starting point to introduce the CDE variation.

The key difference between $\stale$ and the CDE-based model is how the latent embedding $z_{o_t}(\cdot)$ is evolved. Consider Figure~\ref{fig:archs} (bottom), illustrating a CDE-based architecture, and the following accompanying equations, hereby termed the $\cde$ model:
\begin{subequations}
\begin{align}
    z_{s_t} &:= F_s(s_t), \quad z_{o_t}(0) := F_o(z_{s_t}, x_t(0)) \\
    dz_{o_t}^\tau &:= f_o(z_{s_t}, z_{o_t}^\tau)\,\, d\bar{x}_t^\tau, \quad \tau \in [0, T) \\
    u_t(\tau) &:= F_u(z_{o_t}(\tau)),\quad \tau \in [0, T).
\end{align}
\label{cde_model}
\end{subequations}
Notice that the initial action $u_t(0)$ is generated in the same way as $\stale$, as this is fundamentally constrained by the available observations at that instant, i.e., $(s_t, x_t(0))$. For $\tau >0$ however, $\cde$ evolves the multi-sensor latent embedding $z_{o_t}(\cdot)$ via a neural CDE that is conditioned on the seen image (via $z_{s_t}$), and \emph{driven} by the incoming stream of higher-frequency observations $x_t(\cdot)$. Decoding this embedding at any intermediate $\tau$ into a control action is done via the same decoder structure as $\stale$, i.e., via $F_u$.

Some remarks are in order. The $\stale$ model may be seen as an \emph{explicit} map between the incoming observations and the control action, in that it processes the observations in a stateless (\ie non-temporal) manner. In the experiments, we will present two other baseline models that do incorporate some notion of state and temporal reasoning. $\cde$ however may be seen as an \emph{implicit} map that better captures the \emph{functional} nature of HCT policies, by explicitly reasoning about the temporal structure of the multi-sensory observations. Such a representation has two key consequences. First, by the universality of the CDE representation (see~\cite{kidger2020neural}), such a model subsumes ODE-based parameterizations of the form:
\[
    \dfrac{d z_{o_t}(\tau)}{d \tau} = f(z_{s_t}, z_{o_t}(\tau), \bar{x}_t(\tau)),
\]
and therefore, generalizes \rev{continuous-time RNNs} and deep DMP-based models referenced at the beginning of this work. \rev{Second, by working directly within the functional space, i.e., by seeking maps from $(s_t, x_t)$ to the control function $u_t$, we avoid having to force discrete-time latent prediction models such as RNNs to handle multiple asynchronous and irregular time-series, e.g., by alternating Neural ODE solves and RNN jumps~\cite{rubanova2019latent}}. Third, the CDE itself may be interpreted as capturing the higher-frequency/higher-fidelity environment dynamics taking place in-between successive images, thereby imbuing the policy network with a model-based inductive bias that is crucial for reasoning over the space of trajectories.

\subsection{Implementation and Imitation Learning}\label{sec:bc}

Notice that the $\cde$ model in~\eqref{cde_model} requires integrating against a $\mathcal{C}^2$ interpolant of the raw time-series $\bm{x}_t$. While natural cubic splines fulfill this criteria, they induce a non-causal dependence on the time-series $\bm{x}_t$ since the coefficients of the spline depend upon the entire set $\bm{x}_t$. Following~\cite{morrill2021neural}, we instead implement the ``discretely-online" strategy whereby we first assume that control is output at a pre-determined frequency that is a multiple of the image-arrival frequency. Thus, for $\tau \in [0, T]$ between discrete time-steps $\{t, t+1\}$, we are required to output $T$ actions at times $\{\tau_i\}_{i=0}^{T-1}$, where without loss of generality, we take $\tau_{i+1}-\tau_{i} = 1$ and $\tau_0 = 0$. The CDE model in~\eqref{cde_model} is then evolved in intervals of $1$, where for generating the control action at time $\tau_i, i>0$, we use the subset of measurements $\bm{x}_t$ lying in-between $[\tau_{i-1}, \tau_i]$ to form our cubic spline. Thus, the model remains causal at the discrete control frequency. If truly continuous-time causal control is required, one may evolve the CDE using a zero-order-hold interpolant of the measurements $\bm{x}_t$.

While either RL or IL may be used to train HCT policies, in this work, we focus on the imitation case - specifically, behavior cloning (BC). To this end, we assume the expert's dataset is a collection of observation-action tuples $\{(o_t, \bm{u}_t)\}_t \sim \mathcal{P}_e$, where $o_t$ is the tuple $(s_t, \bm{x}_t)$, and $\bm{u}_t := \{u_t(0), u_t(1), \ldots, u_t(T-1)\}$ are the sampled expert actions. Let $\theta$ represent the set of all trainable parameters (across the encoder, decoder, and neural CDE), and let $u_{t,\theta}(\cdot | o_t): [0, T) \rightarrow \U$ denote the control function generated by the HCT policy, as a function of the observations $o_t$. BC training is performed by minimizing the objective:
\begin{equation}
    \min_{\theta} \ \hat{\mathbb{E}}_{(o_t, \bm{u}_t) \sim \mathcal{P}_e}\ \  \dfrac{1}{T}\int_{0}^{T-1} l(u_{t,\theta}(\tau | o_t), \hat{u}_t(\tau))\ d\tau,
\label{bc-loss}
\end{equation}
where $l(\cdot, \cdot): \U \times \U \rightarrow \reals_{\geq 0}$ is a non-negative cost function penalizing the difference between the predicted and actual control output, and $\hat{u}_t(\cdot)$ is a piecewise $\mathcal{C}^1$ interpolant of the sampled signal $\bm{u}_t$. \rev{We remark that since an HCT policy models control \emph{functions}, the loss is written as an integral rather than a sum over the finite set of observed control values.}

\section{Experiment Setup and Hypotheses}

We performed several variations of BC experiments on data collected by expert policies (both scripted and pre-trained with privileged obeservations) on a variety of environments. In this section we outline the environment and tasks, types of experiments, and comparative baseline models.

\subsection{Environments}

\begin{figure}[h]
\centering
\begin{subfigure}[b]{0.23\textwidth}
 \centering
 \includegraphics[width=\textwidth]{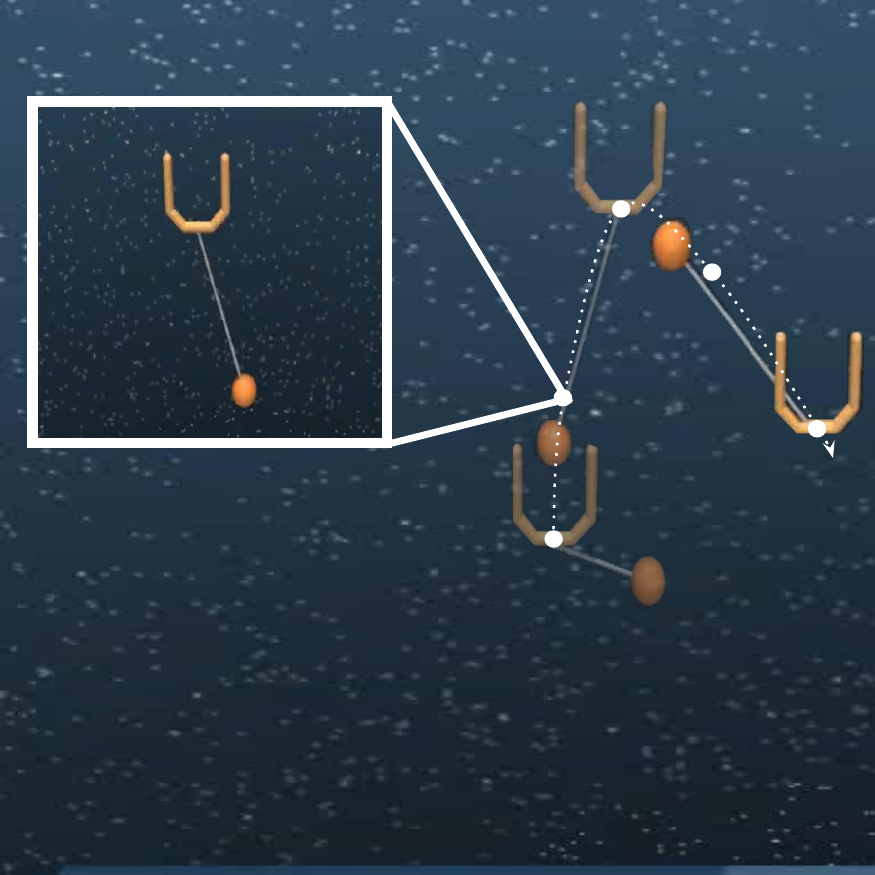}
\end{subfigure}
\begin{subfigure}[b]{0.23\textwidth}
 \centering
 \includegraphics[width=\textwidth]{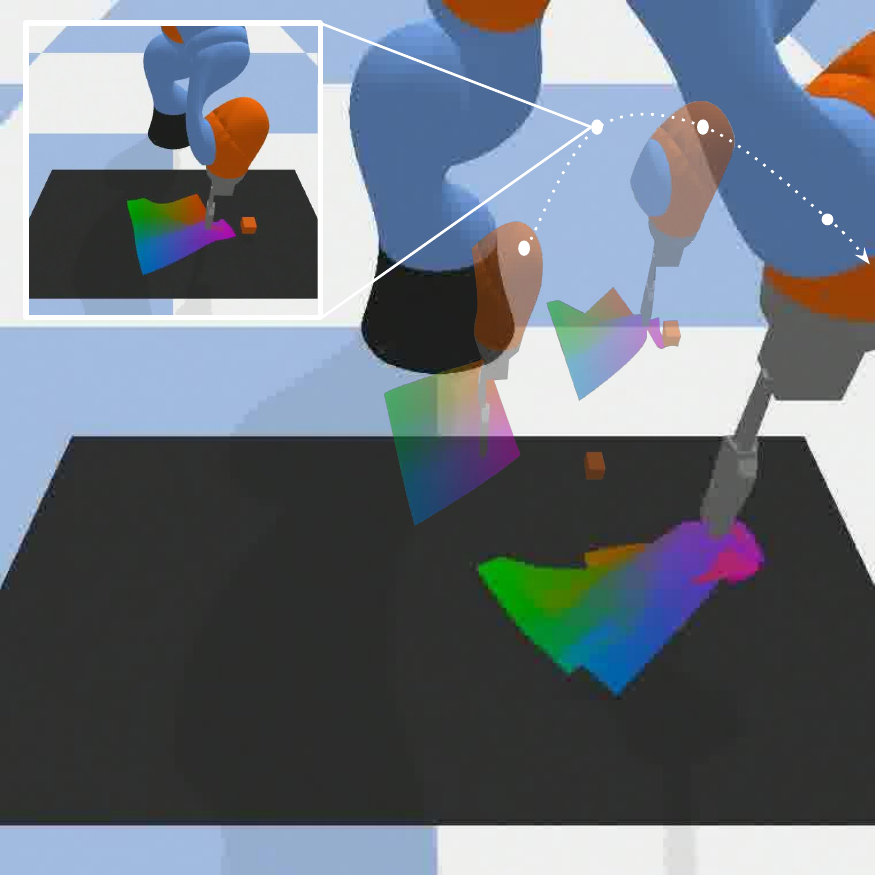}
\end{subfigure}
\caption{\footnotesize{Example Timelapse for evaluation environments; \emph{Left}: Ball-In-Cup, \emph{Right}: Cloth Covering.}}
\label{fig:envs}
\vspace{-3mm}
\end{figure}

Figure~\ref{fig:envs} depicts the two evaluation environments: Cloth-Covering (CC): where the objective is to pick-up and drape a towel over an object (the locations of both items are randomized upon initialization), and Ball-In-Cup (BiC): consisting of an actuated cup in the vertical plane attempting to swing and catch a ball attached via an elastic cable~\cite{tassa2018deepmind}. In addition to the incoming image stream $s_t$, the higher-frequency observations $x_t$ correspond to the robot proprioceptive sensors for the CC environment and cup position and velocity for BiC. By denying explicit observation of the ball position, we are forcing the policy to implicitly deduce and forward predict this information from the image stream.

We also study a harder variant of the BiC environment - variable BiC (vBiC), where the higher-frequency observation is reduced to a 3-dof force-torque sensor collocated at the cup position, and the ball mass and maximum length of the string are randomized at the beginning of each episode to lie within a $\pm 1/3$ range of their nominal values. We study this variation to examine the effect the nature of the higher-frequency observations may have on the learnability of the policy. Please see Appendix~\ref{app:envs}\footnote{All appendices referenced herein may be found within the online version of this work~\cite{SinghRamirez2022}.} for additional details on the environments.

\subsection{BC Evaluations: Robustness to Latency and Drops}

For each environment, we specified a nominal value of $T$ (the time in-between successive image experiments), denoted as $T_0$, and performed BC training and evaluation as outlined in Section~\ref{sec:bc}. In addition, we introduce two new variations custom to the HCT control setting: (i) {\it Dropped-BC}, and (ii) {\it Throttled-BC}. \emph{Dropped-BC} emulates dropped image sensor observations, whereby the policy is forced to rely upon the last image saved in memory if the expected incoming image at the current time index $t$ is lost. We quantize the difficulty of this BC variation via a Bernoulli \emph{drop probability} $p_d \in [0, 1)$, specifying the probability of a missed image at any time index $t$. To perform this experiment, we took the nominal models trained on un-corrupted data (i.e., $p_d = 0$) and fine-tuned on corrupted data at $p_d = 0.1$. We then evaluated the fine-tuned policies at a \emph{range} of drop probabilities within $[0, 0.5]$.

Within {\it Throttled-BC}, the time in-between successive images (equivalently, the number of control actions to be generated between two successive images), i.e., $T$ is varied. We hypothesize that for tasks where the policy may be sensitive to the higher frequency observations, increasing $T$ should correlate with an increase in difficulty, necessitating non-trivial latent prediction. Thus, \emph{Throttled-BC} may be seen as a \rev{deterministic} limit of \emph{Dropped-BC}. To perform this experiment, we fine-tuned each model on data corresponding to different values of $T$ by warmstarting\footnote{Note that no additional data was collected - we simply paired an image $s_t$ with a longer sequence of observations $\bm{x}_t$ and actions $\bm{u}_t$, thus effectively \emph{reducing} the ``dataset size" by the factor $T/T_0$.} with parameters trained at $T=T_0$.

\subsection{Baselines}

In addition to the $\stale$ model given by~\eqref{stale_model}, we compare against two additional baselines that incorporate some temporal structure into the policy. The first, termed $\mastale$ corresponds to a variation of $\stale$ whereby one leverages an exponential-moving-average filter of the higher-frequency observations $\bm{x}_t$ in place of the direct feedthrough of the interpolant $\hat{x}_t(\cdot)$. This imbues the $\stale$ model with some stateful reasoning over the time-series $\bm{x}_t$. The final baseline corresponds to the Neural Dynamic Policies ($\ndp$) model introduced in~\cite{bahl2020neural}, whereby the observations $(s_t, x_t(0))$ are encoded into a set of parameters for a DMP that governs the \emph{open-loop} evolution of $u_t(\cdot)$. Please see Appendix~\ref{app:base} for the relevant equations and additional architecture details.

\section{Results}

\begin{figure}[h]
\centering
\begin{subfigure}[b]{0.5\textwidth}
 \centering
 \includegraphics[width=\textwidth]{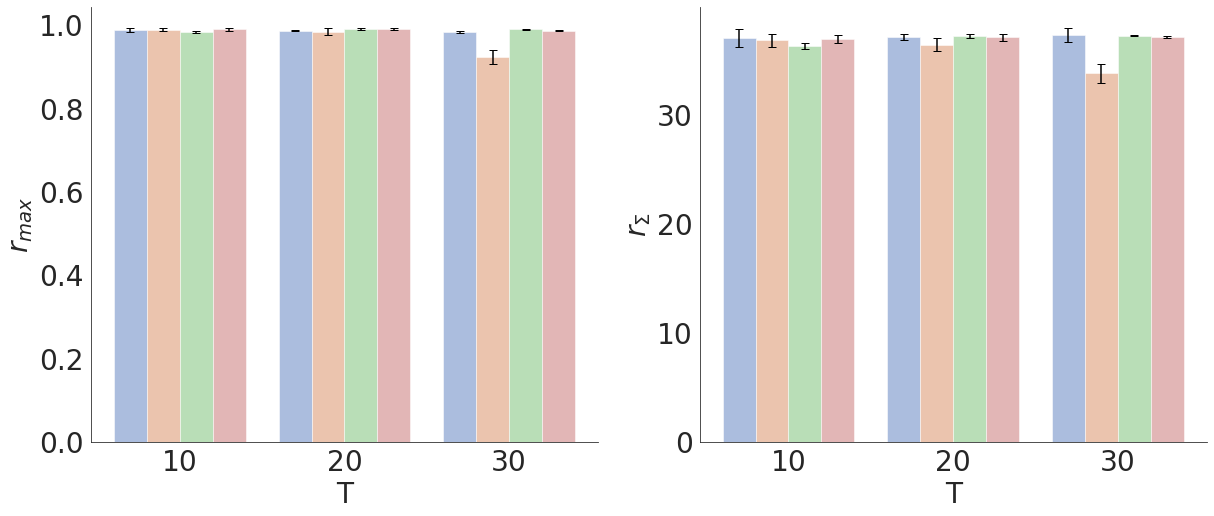}
\end{subfigure}
\begin{subfigure}[b]{0.5\textwidth}
 \centering
 \includegraphics[width=\textwidth]{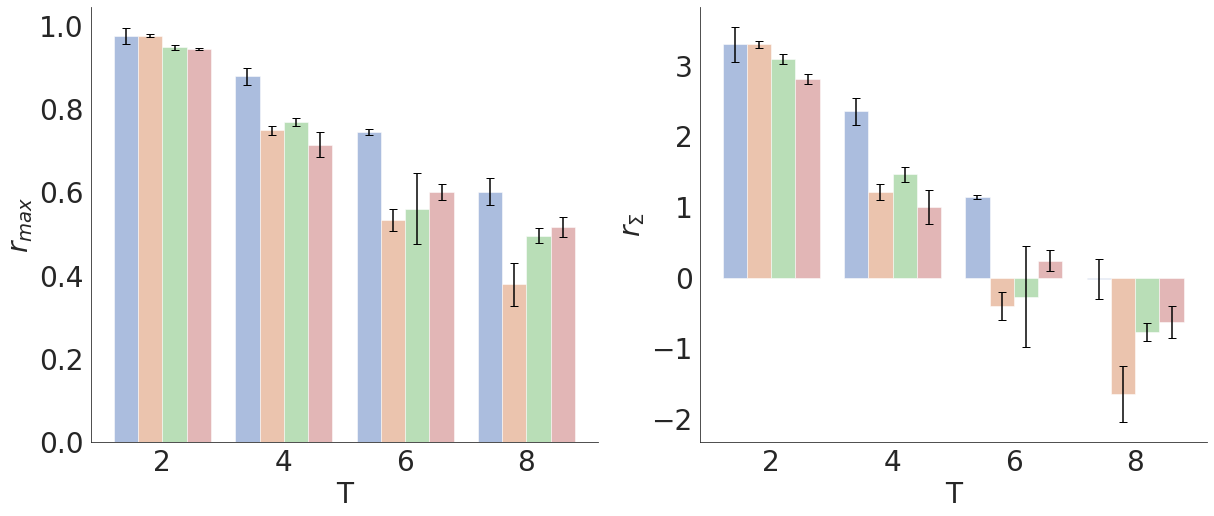}
\end{subfigure}
\begin{subfigure}[b]{0.5\textwidth}
 \centering
 \includegraphics[width=\textwidth]{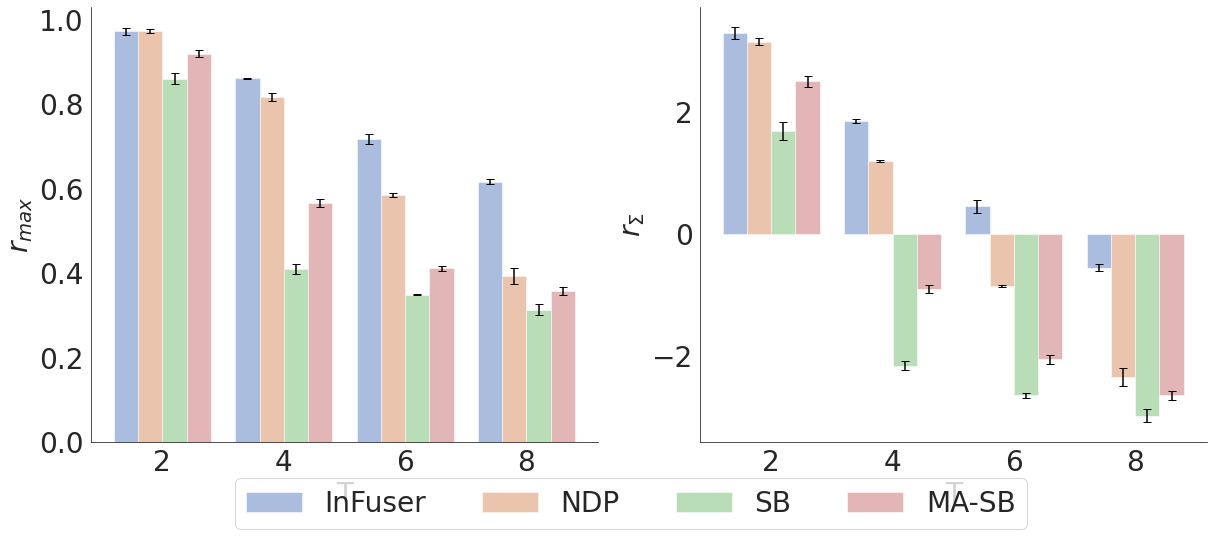}
\end{subfigure}
\caption{\footnotesize{Throttled-BC: Average (over 15 rollouts) $\rmax$ (\emph{left}), and $\rtot$ (\emph{right}), for varying values of $T$ for cloth-covering (\emph{top}), ball-in-cup (\emph{middle}), and variable ball-in-cup (\emph{bottom}). $\rmax=1$ indicates task success while larger values of $\rtot$ indicate a faster time to completion. The presented values and $\pm 1\sigma$ ranges are computed across 3 independent seeds.}}
\label{fig:bc-results}
\vspace{-3mm}
\end{figure}

We present two key metrics for comparison: $\rmax$: the maximum reward over an episode and $\rtot$: the total accumulated reward over an episode. For all environments, $\rmax = 1$ is equivalent to ``task success" (i.e., object fully covered by cloth or ball caught in cup). The total reward measures efficiency, i.e., a larger $\rtot$ indicates either more time-steps spent with the object fully covered or less time to catch the ball. Figure~\ref{fig:bc-results} summarizes the \emph{Throttled-BC} performance of all four models across the three environments for varying values of $T$. The left-most column in each subfigure corresponds to the $T=T_0$ nominal setting, while the remaining columns correspond to various levels of image throttling.

On the \emph{Throttled-BC} results, we first highlight that all model architectures are competitive at the lowest values of $T$, indicating that all policy architectures have the requisite capacity to solve the task under ``nearly" MDP conditions. However, there is considerable variability in performance as the image arrival rate is throttled down (i.e., $T$ is increased), particularly for the BiC environments. 

For the CC task, image throttling does not appear to cause any significant degradation in performance. This is likely due to the quasi-static nature of the task. A single snapshot of the scene is sufficient to execute a fairly long open-loop sequence of actions (i.e., the $\ndp$ model). Further, the task is not overly sensitive to the higher-frequency observations. Thus, there is no discernible advantage to the CDE method of incorporating these observations, as compared with the explicit models $\stale$ and $\mastale$.

In contrast, for both the standard and variable BiC environments, $\cde$ outperforms all other models in both metrics as $T$ is increased. The separation is more apparent for the harder vBiC variant where the higher-frequency observation is just the force-torque sensor. \rev{Indeed, ablations within this environment demonstrated that the change in the higher-frequency sensor is the primary source of the larger spread in performance, rather than the variability of the inertial properties.} The $\cde$ architecture's stateful representation is particularly useful in this setting for performing the latent conversion to proprioceptive-level information.
\begin{figure}[h]
\centering
\begin{subfigure}[t]{0.24\textwidth}
 \centering
 \includegraphics[width=\textwidth]{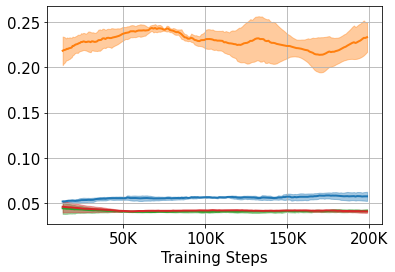}
\end{subfigure}\\
\begin{subfigure}[b]{0.23\textwidth}
 \centering
 \includegraphics[width=\textwidth]{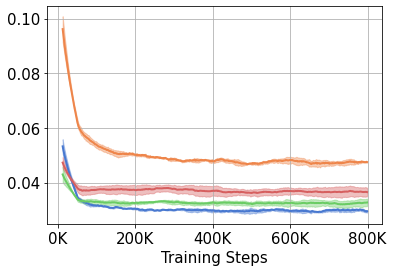}
\end{subfigure}
\begin{subfigure}[b]{0.23\textwidth}
 \centering
 \includegraphics[width=\textwidth]{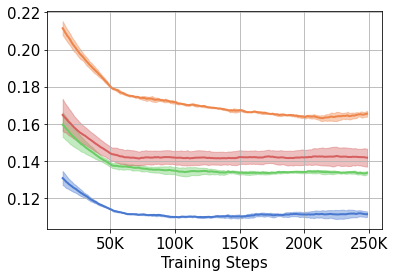}
\end{subfigure}
\caption{\footnotesize{Throttled-BC test loss vs training steps for cloth-covering at $T=20$ (\emph{top}), ball-in-cup at $T=4$ (\emph{bottom-left}), and variable ball-in-cup at $T=4$ (\emph{bottom-right}).}}
\label{fig:bc-loss}
\end{figure}

A plausible justification for the observed trends is that the BiC tasks are a lot more dynamic in nature, necessitating non-trivial latent predictions to generate good action sequences. \rev{Quantifying ``good" however is a rather subtle discussion. Figure~\ref{fig:bc-loss} plots the empirical BC loss defined in~\eqref{bc-loss} on a held-out test set, for each environment at a value of $T$ larger than $T_0$. We observe that while the $\cde$ architecture seems to exhibit a correlation between the test BC loss and closed-loop reward performance, this trend does not generalize to the other models. That is, lower test error does not necessarily imply better closed-loop reward performance, indicating the need for further investigation into the closed-loop \emph{robustness} of these HCT policies. We provide the remaining test error plots in Appendix~\ref{app:results}, further corroborating this observation.}

\begin{figure}[h]
\centering
\begin{subfigure}[b]{0.5\textwidth}
 \centering
 \includegraphics[width=\textwidth]{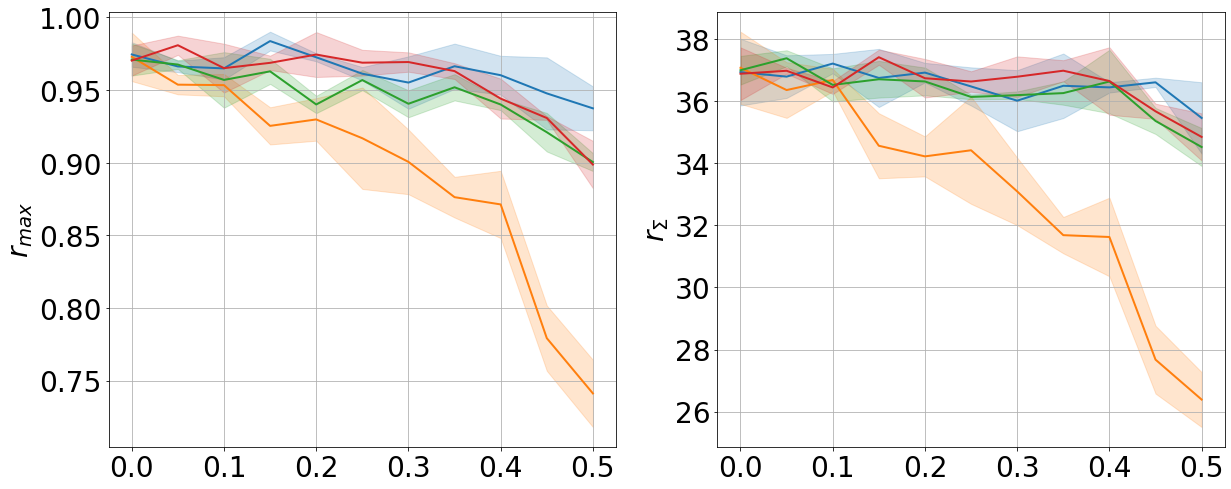}
\end{subfigure}
\begin{subfigure}[b]{0.5\textwidth}
 \centering
 \includegraphics[width=\textwidth]{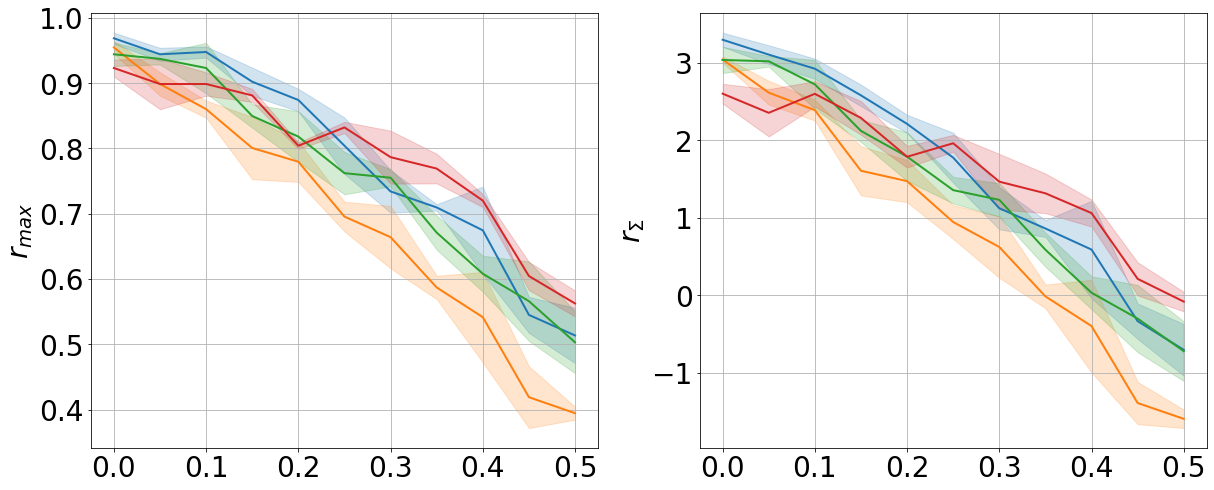}
\end{subfigure}
\begin{subfigure}[b]{0.5\textwidth}
 \centering
 \includegraphics[width=\textwidth]{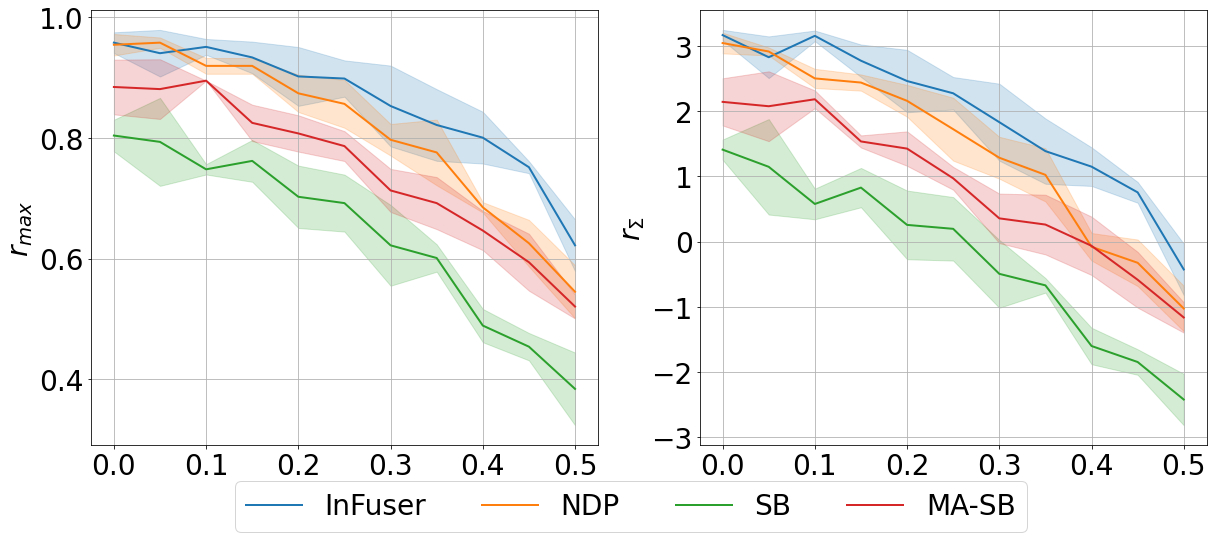}
\end{subfigure}
\caption{\footnotesize{Dropped-BC: Average (over 100 rollouts) $\rmax$ (\emph{left}), and $\rtot$ (\emph{right}), for varying values of $p_d$ (image-drop probability) for cloth-covering (\emph{top}), ball-in-cup (\emph{middle}), and variable ball-in-cup (\emph{bottom}). The parameter $T$ is fixed at the lowest setting from the Throttled-BC experiments. The shaded $\pm 1 \sigma$ range is computed over three seeds.}}
\label{fig:bc-drop}
\vspace{-3mm}
\end{figure}

The aforementioned trends are further reflected in the Dropped-BC experiments (see Figure~\ref{fig:bc-drop}). For the CC task, all models except $\ndp$ remain roughly on par with the ``no-dropping" baseline performance, even as the probability of dropped images $p_d$ is increased, thereby reinforcing the quasi-static nature of the task. For the BiC tasks however, all models degrade in performance as $p_d$ is increased, with $\cde$ being the most performant in the vBiC variation, while being on par with the other models in the standard variation. Once again, this highlights the higher level of requisite agility for the BiC tasks and the inherent robustness of the $\cde$ policy.

\section{Conclusions and Future Work}

In this work we introduced $\cde$, a policy architecture for seamlessly (In)tegrating and (Fus)ing multi-sensory multi-frequency observations for continuous-time control within dynamic environments. The key insight is to treat multi-sensory fusion and temporal abstractions within a unified framework using controlled differential equations. Our CDE-based architecture evolves a continuous-time multi-sensory latent embedding that is conditioned on images, and driven by higher-frequency observations. The resulting model outperforms state-of-the-art multi-sensory fusion architectures and DMP-based policies, particularly when deployed on tasks that demand a non-trivial level of agility.

Our contributions enable several exciting future directions for research. First, although this work only studies the behavior cloning setting, one may easily incorporate this architecture within hybrid continuous-time reinforcement learning algorithms~\cite{xiao2019thinking}. Second, recent work~\cite{florence2022implicit} has demonstrated the importance of multi-modal representations for BC-trained policies. A promising direction therefore is to infuse multi-modality within the CDE architecture to broaden the class of achievable tasks, such as those from the manipulation domain. Third, the proposed CDE model naturally enables deployment on platforms where the speed of image processing may otherwise be a key bottleneck in realizing truly agile continuous-time control.


\bibliographystyle{IEEEtran}
\bibliography{root.bbl}

\begin{thebibliography}{10}
\providecommand{\url}[1]{#1}
\csname url@rmstyle\endcsname
\providecommand{\newblock}{\relax}
\providecommand{\bibinfo}[2]{#2}
\providecommand\BIBentrySTDinterwordspacing{\spaceskip=0pt\relax}
\providecommand\BIBentryALTinterwordstretchfactor{4}
\providecommand\BIBentryALTinterwordspacing{\spaceskip=\fontdimen2\font plus
\BIBentryALTinterwordstretchfactor\fontdimen3\font minus
  \fontdimen4\font\relax}
\providecommand\BIBforeignlanguage[2]{{%
\expandafter\ifx\csname l@#1\endcsname\relax
\typeout{** WARNING: IEEEtran.bst: No hyphenation pattern has been}%
\typeout{** loaded for the language `#1'. Using the pattern for}%
\typeout{** the default language instead.}%
\else
\language=\csname l@#1\endcsname
\fi
#2}}

\bibitem{kidger2020neural}
P.~Kidger, J.~Morrill, J.~Foster, and T.~Lyons, ``Neural controlled
  differential equations for irregular time series,'' in \emph{Advances in
  Neural Information Processing Systems}, 2020.

\bibitem{chen2018neural}
R.~T. Chen, Y.~Rubanova, J.~Bettencourt, and D.~Duvenaud, ``Neural ordinary
  differential equations,'' \emph{arXiv preprint arXiv:1806.07366}, 2018.

\bibitem{lee2019making}
M.~A. Lee, Y.~Zhu, K.~Srinivasan, P.~Shah, S.~Savarese, L.~Fei-Fei, A.~Garg,
  and J.~Bohg, ``Making sense of vision and touch: Self-supervised learning of
  multimodal representations for contact-rich tasks,'' in \emph{2019
  International Conference on Robotics and Automation (ICRA)}.\hskip 1em plus
  0.5em minus 0.4em\relax IEEE, 2019, pp. 8943--8950.

\bibitem{li2019connecting}
Y.~Li, J.-Y. Zhu, R.~Tedrake, and A.~Torralba, ``Connecting touch and vision
  via cross-modal prediction,'' in \emph{Proceedings of the IEEE/CVF Conference
  on Computer Vision and Pattern Recognition}, 2019, pp. 10\,609--10\,618.

\bibitem{lee2021detect}
M.~A. Lee, M.~Tan, Y.~Zhu, and J.~Bohg, ``Detect, reject, correct: Crossmodal
  compensation of corrupted sensors,'' in \emph{2021 IEEE International
  Conference on Robotics and Automation (ICRA)}.\hskip 1em plus 0.5em minus
  0.4em\relax IEEE, 2021, pp. 909--916.

\bibitem{liu2020understanding}
Y.~Liu, D.~Romeres, D.~K. Jha, and D.~Nikovski, ``Understanding multi-modal
  perception using behavioral cloning for peg-in-a-hole insertion tasks,''
  \emph{arXiv preprint arXiv:2007.11646}, 2020.

\bibitem{gao2021objectfolder}
R.~Gao, Y.-Y. Chang, S.~Mall, L.~Fei-Fei, and J.~Wu, ``Objectfolder: A dataset
  of objects with implicit visual, auditory, and tactile representations,''
  \emph{arXiv preprint arXiv:2109.07991}, 2021.

\bibitem{yang2021learning}
R.~Yang, M.~Zhang, N.~Hansen, H.~Xu, and X.~Wang, ``Learning vision-guided
  quadrupedal locomotion end-to-end with cross-modal transformers,''
  \emph{arXiv preprint arXiv:2107.03996}, 2021.

\bibitem{calandra2018more}
R.~Calandra, A.~Owens, D.~Jayaraman, J.~Lin, W.~Yuan, J.~Malik, E.~H. Adelson,
  and S.~Levine, ``More than a feeling: Learning to grasp and regrasp using
  vision and touch,'' \emph{IEEE Robotics and Automation Letters}, vol.~3,
  no.~4, pp. 3300--3307, 2018.

\bibitem{hu2021unit}
R.~Hu and A.~Singh, ``{UniT}: Multimodal multitask learning with a unified
  transformer,'' \emph{arXiv preprint arXiv:2102.10772}, 2021.

\bibitem{sung2017deep}
J.~Sung, I.~Lenz, and A.~Saxena, ``Deep multimodal embedding: Manipulating
  novel objects with point-clouds, language and trajectories,'' in \emph{2017
  IEEE International Conference on Robotics and Automation (ICRA)}.\hskip 1em
  plus 0.5em minus 0.4em\relax IEEE, 2017, pp. 2794--2801.

\bibitem{de2018integrating}
T.~De~Bruin, J.~Kober, K.~Tuyls, and R.~Babu{\v{s}}ka, ``Integrating state
  representation learning into deep reinforcement learning,'' \emph{IEEE
  Robotics and Automation Letters}, vol.~3, no.~3, pp. 1394--1401, 2018.

\bibitem{lesort2018state}
T.~Lesort, N.~D{\'\i}az-Rodr{\'\i}guez, J.-F. Goudou, and D.~Filliat, ``State
  representation learning for control: An overview,'' \emph{Neural Networks},
  vol. 108, pp. 379--392, 2018.

\bibitem{saveriano2021dynamic}
M.~Saveriano, F.~J. Abu-Dakka, A.~Kramberger, and L.~Peternel, ``Dynamic
  movement primitives in robotics: A tutorial survey,'' \emph{arXiv preprint
  arXiv:2102.03861}, 2021.

\bibitem{khansari2011learning}
S.~M. Khansari-Zadeh and A.~Billard, ``Learning stable nonlinear dynamical
  systems with {Gaussian} mixture models,'' \emph{IEEE Transactions on
  Robotics}, vol.~27, no.~5, pp. 943--957, 2011.

\bibitem{khansari2014learning}
------, ``Learning control {Lyapunov} function to ensure stability of dynamical
  system-based robot reaching motions,'' \emph{Robotics and Autonomous
  Systems}, vol.~62, no.~6, pp. 752--765, 2014.

\bibitem{khader2021learning}
S.~A. Khader, H.~Yin, P.~Falco, and D.~Kragic, ``Learning stable
  normalizing-flow control for robotic manipulation,'' in \emph{2021 IEEE
  International Conference on Robotics and Automation (ICRA)}.\hskip 1em plus
  0.5em minus 0.4em\relax IEEE, 2021, pp. 1644--1650.

\bibitem{singh2021learning}
S.~Singh, S.~M. Richards, V.~Sindhwani, J.-J.~E. Slotine, and M.~Pavone,
  ``Learning stabilizable nonlinear dynamics with contraction-based
  regularization,'' \emph{The International Journal of Robotics Research},
  vol.~40, no. 10-11, pp. 1123--1150, 2021.

\bibitem{sindhwani2018learning}
V.~Sindhwani, S.~Tu, and M.~Khansari, ``Learning contracting vector fields for
  stable imitation learning,'' \emph{arXiv preprint arXiv:1804.04878}, 2018.

\bibitem{khadir2019teleoperator}
B.~E. Khadir, J.~Varley, and V.~Sindhwani, ``Teleoperator imitation with
  continuous-time safety,'' \emph{arXiv preprint arXiv:1905.09499}, 2019.

\bibitem{rai2017learning}
A.~Rai, G.~Sutanto, S.~Schaal, and F.~Meier, ``Learning feedback terms for
  reactive planning and control,'' in \emph{2017 IEEE International Conference
  on Robotics and Automation (ICRA)}.\hskip 1em plus 0.5em minus 0.4em\relax
  IEEE, 2017, pp. 2184--2191.

\bibitem{chebotar2014learning}
Y.~Chebotar, O.~Kroemer, and J.~Peters, ``Learning robot tactile sensing for
  object manipulation,'' in \emph{2014 IEEE/RSJ International Conference on
  Intelligent Robots and Systems}.\hskip 1em plus 0.5em minus 0.4em\relax IEEE,
  2014, pp. 3368--3375.

\bibitem{daniel2012hierarchical}
C.~Daniel, G.~Neumann, and J.~Peters, ``Hierarchical relative entropy policy
  search,'' in \emph{Artificial Intelligence and Statistics}.\hskip 1em plus
  0.5em minus 0.4em\relax PMLR, 2012, pp. 273--281.

\bibitem{stulp2012reinforcement}
F.~Stulp, E.~A. Theodorou, and S.~Schaal, ``Reinforcement learning with
  sequences of motion primitives for robust manipulation,'' \emph{IEEE
  Transactions on robotics}, vol.~28, no.~6, pp. 1360--1370, 2012.

\bibitem{parisi2015reinforcement}
S.~Parisi, H.~Abdulsamad, A.~Paraschos, C.~Daniel, and J.~Peters,
  ``Reinforcement learning vs human programming in tetherball robot games,'' in
  \emph{2015 IEEE/RSJ International Conference on Intelligent Robots and
  Systems (IROS)}.\hskip 1em plus 0.5em minus 0.4em\relax IEEE, 2015, pp.
  6428--6434.

\bibitem{sutton1999between}
R.~S. Sutton, D.~Precup, and S.~Singh, ``Between {MDPs} and {semi-MDPs}: A
  framework for temporal abstraction in reinforcement learning,''
  \emph{Artificial intelligence}, vol. 112, no. 1-2, pp. 181--211, 1999.

\bibitem{peters2008reinforcement}
J.~Peters and S.~Schaal, ``Reinforcement learning of motor skills with policy
  gradients,'' \emph{Neural networks}, vol.~21, no.~4, pp. 682--697, 2008.

\bibitem{bahl2020neural}
S.~Bahl, M.~Mukadam, A.~Gupta, and D.~Pathak, ``Neural dynamic policies for
  end-to-end sensorimotor learning,'' \emph{arXiv preprint arXiv:2012.02788},
  2020.

\bibitem{bahl2021hierarchical}
S.~Bahl, A.~Gupta, and D.~Pathak, ``Hierarchical neural dynamic policies,''
  \emph{arXiv preprint arXiv:2107.05627}, 2021.

\bibitem{narita2021policy}
T.~Narita and O.~Kroemer, ``Policy blending and recombination for multimodal
  contact-rich tasks,'' \emph{IEEE Robotics and Automation Letters}, vol.~6,
  no.~2, pp. 2721--2728, 2021.

\bibitem{escontrela2020zero}
A.~Escontrela, G.~Yu, P.~Xu, A.~Iscen, and J.~Tan, ``Zero-shot terrain
  generalization for visual locomotion policies,'' \emph{arXiv preprint
  arXiv:2011.05513}, 2020.

\bibitem{florence2022implicit}
P.~Florence, C.~Lynch, A.~Zeng, O.~A. Ramirez, A.~Wahid, L.~Downs, A.~Wong,
  J.~Lee, I.~Mordatch, and J.~Tompson, ``Implicit behavioral cloning,'' in
  \emph{Conference on Robot Learning}, 2022, pp. 158--168.

\bibitem{zeng2020transporter}
A.~Zeng, P.~Florence, J.~Tompson, S.~Welker, J.~Chien, M.~Attarian,
  T.~Armstrong, I.~Krasin, D.~Duong, V.~Sindhwani, \emph{et~al.}, ``Transporter
  networks: Rearranging the visual world for robotic manipulation,''
  \emph{arXiv preprint arXiv:2010.14406}, 2020.

\bibitem{jang2022bc}
E.~Jang, A.~Irpan, M.~Khansari, D.~Kappler, F.~Ebert, C.~Lynch, S.~Levine, and
  C.~Finn, ``{BC-Z}: Zero-shot task generalization with robotic imitation
  learning,'' in \emph{Conference on Robot Learning}.\hskip 1em plus 0.5em
  minus 0.4em\relax PMLR, 2022, pp. 991--1002.

\bibitem{oksendal2013stochastic}
B.~{\O}ksendal, \emph{Stochastic differential equations: an introduction with
  applications}.\hskip 1em plus 0.5em minus 0.4em\relax Springer Science \&
  Business Media, 2013.

\bibitem{lyons2007differential}
T.~J. Lyons, M.~Caruana, and T.~L{\'e}vy, \emph{Differential equations driven
  by rough paths}.\hskip 1em plus 0.5em minus 0.4em\relax Springer, 2007.

\bibitem{rubanova2019latent}
Y.~Rubanova, R.~T. Chen, and D.~K. Duvenaud, ``Latent ordinary differential
  equations for irregularly-sampled time series,'' in \emph{Advances in neural
  information processing systems}, 2019.

\bibitem{morrill2021neural}
J.~Morrill, P.~Kidger, L.~Yang, and T.~Lyons, ``Neural controlled differential
  equations for online prediction tasks,'' \emph{arXiv preprint
  arXiv:2106.11028}, 2021.

\bibitem{tassa2018deepmind}
Y.~Tassa, Y.~Doron, A.~Muldal, T.~Erez, Y.~Li, D.~d.~L. Casas, D.~Budden,
  A.~Abdolmaleki, J.~Merel, A.~Lefrancq, \emph{et~al.}, ``Deepmind control
  suite,'' \emph{arXiv preprint arXiv:1801.00690}, 2018.

\bibitem{SinghRamirez2022}
S.~Singh, F.~M. Ramirez, J.~Varley, A.~Zeng, and V.~Sindhwani, ``Multiscale
  sensor fusion and continuous control with neural {CDEs},'' \emph{To be
  uploaded on arXiv.}, 2022.

\bibitem{xiao2019thinking}
T.~Xiao, E.~Jang, D.~Kalashnikov, S.~Levine, J.~Ibarz, K.~Hausman, and
  A.~Herzog, ``Thinking while moving: Deep reinforcement learning with
  concurrent control,'' in \emph{International Conference on Learning
  Representations}, 2020.

\bibitem{choromanski2020provably}
K.~Choromanski, A.~Pacchiano, J.~Parker-Holder, Y.~Tang, D.~Jain, Y.~Yang,
  A.~Iscen, J.~Hsu, and V.~Sindhwani, ``Provably robust blackbox optimization
  for reinforcement learning,'' in \emph{Conference on Robot Learning}, 2020,
  pp. 683--696.

\bibitem{kingma2014adam}
D.~P. Kingma and J.~Ba, ``Adam: A method for stochastic optimization,''
  \emph{arXiv preprint arXiv:1412.6980}, 2014.

\bibitem{smith2017cyclical}
L.~N. Smith, ``Cyclical learning rates for training neural networks,'' in
  \emph{2017 IEEE winter conference on applications of computer vision
  (WACV)}.\hskip 1em plus 0.5em minus 0.4em\relax IEEE, 2017, pp. 464--472.

\bibitem{jax2018github}
\BIBentryALTinterwordspacing
J.~Bradbury, R.~Frostig, P.~Hawkins, M.~J. Johnson, C.~Leary, D.~Maclaurin,
  G.~Necula, A.~Paszke, J.~Vander{P}las, S.~Wanderman-{M}ilne, and Q.~Zhang,
  ``{JAX}: composable transformations of {P}ython+{N}um{P}y programs,'' 2018.
  [Online]. Available: \url{http://github.com/google/jax}
\BIBentrySTDinterwordspacing

\bibitem{flax2020github}
\BIBentryALTinterwordspacing
J.~Heek, A.~Levskaya, A.~Oliver, M.~Ritter, B.~Rondepierre, A.~Steiner, and
  M.~van {Z}ee, ``{F}lax: A neural network library and ecosystem for {JAX},''
  2020. [Online]. Available: \url{http://github.com/google/flax}
\BIBentrySTDinterwordspacing

\end{thebibliography}

\iftoggle{arxiv}{
\newpage

\onecolumn

\appendices

\section{Environments}
\label{app:envs}

\subsection{Cloth-Covering}
The set of all higher-frequency observations within $x_t$ include: robot joint-angles and velocities, end-effector Cartesian position, and gripper status (open vs closed), yielding a net higher-frequency observation dimension of $n=30$. The reward function at each step is:
\[
    r_t = 1 - \mathrm{occ}_{\mathrm{ratio}},
\]
where $\mathrm{occ}_{\mathrm{ratio}}$ is the occlusion ratio for the object, computed as the ratio of the object's visible pixel surface area to the object's total pixel surface area. The controller outputs are the change in end-effector Cartesian position (3 dof), and gripper status (1 dof). Note that as the gripper action is a binary variable, the control space dimension is 5, with gripper controls computed as logits.

Expert demonstrations are provided through a scripted policy which, using privileged simulator state regarding cloth and block positions, drives the arm to the center of the cloth, closes the gripper, drives the arm over-top the block position, and opens the gripper. Each episode lasts 120 control steps, corresponding to $120 / T$ seen images.

\subsection{Ball-In-Cup}

The higher-frequency observation is simply the cup 2D position and velocity, i.e., $n=4$, and the reward function is:
\[
    r_t = \begin{cases} 1 \quad &\text{if ball is in cup} \\
    -0.05 \quad &\text{else}.
    \end{cases}.
\]
Further, the environment terminates early after $5$ consecutive steps of $r_t = 1$ (i.e., ball is in the cup); the maximum episode length is 100 control steps. The small negative penalty allow us to measure task efficiency (i.e., how quickly is the catch completed) via the total episode reward $r_{\Sigma}$. The expert data was generated by a separate MLP policy, trained via Blackbox optimization~\cite{choromanski2020provably}, with access to the cup \emph{and} ball position and velocities.

\subsection{Variable Ball-In-Cup}

The variable BiC (vBiC) environment is a harder version of BiC whereby the mass of the ball and the max length of the elastic string are randomized at the beginning of each episode. In particular, we allow both parameters to vary between $\pm 1/3$ of the nominal value, a non-negligible range. To make the task harder, the higher-frequency observation is reduced to just a 3-dof force-torque sensor collocated at the cup position. This variation represents the closest analogue to a human performing the same task, where the set of observations are visual (the RGB camera) and force-feedback. The expert policy for this task was again trained via Blackbox optimization, with access to the cup and ball positions and velocities, as well as the mass of the ball and max length of the string. The maximum episode length is 120 control steps with early termination after $5$ consecutive steps with the ball in the cup.

\section{Models and Architectures}
\label{app:base}

We first provide the relevant equations for the two additional baselines. Within $\mastale$, we replace the second equation in~\eqref{stale_model} with:
\begin{subequations}
\begin{alignat}{2}
    z_{o_t}(\tau) &:= F_o(z_{s_t}, \breve{x}_t(\tau)),\quad &&\tau \in [0, T) \\
    \dfrac{d \breve{x}_t(\tau)}{d \tau} &:= \dfrac{1}{\sigma} (\hat{x}_t(\tau) - \breve{x}_t(\tau)), \quad &&\tau \in [0, T),
\end{alignat}
\label{stale_ma_model}
\end{subequations}
where $\sigma \in \real_{>0}$ is the filter time-constant, set in all experiments to be the timestep in between successive higher-frequency observations. The filter is initialized as $\breve{x}_t(0) = x_t(0)$, and is driven by the cubic interpolant $\hat{x}_t(\cdot)$ of the raw measurements $\bm{x}_t$.

Adapting notation from~\cite{bahl2020neural}, the $\ndp$ model is summarized by the following encoder:
\begin{subequations}
\begin{align}
    z_{s_t} := F_s(s_t), & \quad z_{o_t}^0 := F_o(z_{s_t}, x_t^0) \\
    \begin{bmatrix} u_t(0) \\ \dfrac{d u_t(0)}{d \tau} \end{bmatrix} &:= F_u(z_{o_t}^0) \\
    \theta_{\mathrm{dmp}} := \{g_t, W_t\} &= F_{\mathrm{dmp}}(z_{s_t}, x_t^0),
\end{align}
\label{ndp_enc}
\end{subequations}
and pair of ODEs:
\begin{subequations}
\begin{align}
    \dfrac{d^2 u_t(\tau)}{d \tau^2} &:= \alpha_u \left(\beta (g_t - u_t(\tau)) - \dfrac{d u_t(\tau)}{d \tau}\right) + f_t(\phi_t(\tau)), \quad \tau \in [0, T) \\
    \dfrac{d \phi_t(\tau)}{d \tau} &:= -\alpha_\phi \phi_t(\tau), \quad \tau \in [0, T).
\end{align}
\label{ndp_ode}
\end{subequations}

Here $\phi_t$ is a phase variable, initialized as $\phi_t(0) = 1$, and $\{\alpha_u, \alpha_\phi, \beta\}$ are strictly positive constants. The forcing function $f_t$ takes the form:
\[
    f_t(\phi) = \dfrac{\phi}{\sum_{k=1}^N \psi_k(\phi)} (W_t \psi(\phi)) \circ (g_t - u_t(0)),
\]
where $\psi(\phi):= (\psi_1(\phi), \ldots, \psi_N(\phi)) \in \reals^N$ is a vector of Gaussian radial basis functions evaluated at $\phi$, $N$ is the number of basis functions, and $\circ$ denotes the Hadamard product\footnote{Note that each row of the matrix $W_t$ corresponds to a unique control dimension, and is an $N-$dimensional vector of weights.}. Once again, we see that the computation of $u_t(0)$ takes the same structural form as the other baselines and $\cde$. However, unlike $\cde$ which continuously incorporates the higher-frequency measurements $\bm{x}_t$ along with the fixed image $s_t$, the $\ndp$ generates an \emph{open-loop} control trajectory between $t$ and $t+1$ using only $(s_t, x_t(0))$.

\subsection{Architecture}\label{app:arch}

For fair comparison, all models shared the same image encoder $F_s$ -- a 10-layer Convolutional ResNet with ReLu activations and input-layer BatchNorm. The penultimate embedding is average-pooled, flattened, and passed through a single dense layer. Note that the $\ndp$ encoder also includes the map $F_{\mathrm{dmp}}$, consisting of two 2-layer MLPs mapping $(z_{s_t}, x_t^0)$ to the DMP parameters $g_t$ and $W_t$.

All models also shared the same fusion network $F_o$ -- a 4-layer MLP with ReLu activations. The final embedding is concatenated with the high-frequency observation input into $F_o$ so that $z_{o_t}(\tau)$ includes $\hat{x}_t(\tau)$ (or $\breve{x}_t(\tau)$ for $\mastale$) as a sub-vector. 

Finally, all models shared the same decoder network $F_u$ -- a 4-layer MLP with ReLu activations. Note that the $\ndp$ decoder outputs both $u_t(0)$ and $du_t(0)/d\tau$, while for all other models, the decoder outputs only $u_t(\tau)$.

For the $\cde$ model, the CDE matrix-valued function $f_o$ is a bottleneck 2-layer MLP with ReLu activation for the hidden layer, and $\tanh$ non-linearity for the output layer, as recommended by~\cite{kidger2020neural}. This parameterized function is stacked on top of the matrix $\begin{bmatrix} I_n & \bm{0}_n \end{bmatrix}$ where $I_n$ is the $n\times n$ identity matrix and $\bm{0}_n$ is an $n-$dimensional column vector of zeros. This is done since $z_{o_t}(\tau)$ includes $\hat{x}_t(\tau)$ as a sub-vector.

We also employed an $l_2$ weight regularization penalty for the parameters of $f_o$ to keep the vector field well conditioned for the ODE solver (fixed-step rk4).

\section{Training and Evaluation}

\noindent {\bf Normalization}: The higher-frequency observations $\bm{x}_t$ were normalized for the CC and vBiC tasks to have zero mean and unit variance. Additionally, the actions were also normalized to the range $[-1, 1]$ for the CC task, while the action space was already normalized for the BiC and vBiC tasks to lie within the range $[-1, 1]$.
\medskip

\noindent {\bf Optimizer}: We used the Adam optimizer~\cite{kingma2014adam} for all training and fine-tuning, along with triangular cyclical\footnote{With the exception of the BiC task at $T=T_0$, where we used a larger constant learning rate to avoid premature convergence to bad local minima.} learning rate schedules~\cite{smith2017cyclical}, with a $\max/\min$ ratio of $5$ and cycle length of $8 \times \# \text{steps per epoch}$. The maximum learning rate was decayed by a factor of $5$ if the training loss was observed to stall for $50$ epochs, and training was terminated early once the stall lasted for $100$ epochs. The initial minimum (constant for BiC) learning rates for the $T=T_0$ cycles were: $10^{-3}$ (BiC), $10^{-4}$ (CC), and $5\cdot 10^{-4}$ (vBiC). For all fine-tuning jobs, i.e., Throttled-BC ($T > T_0$) and Dropped-BC ($T = T_0, p_d = 0.1$), we only used cyclical learning rates with initial minimum cycle values: $10^{-4}$ (BiC), $5\cdot 10^{-5}$ (CC), and $10^{-4}$ (vBiC). The batch sizes were $64$ (CC) and $128$ (BiC and vBiC).
\medskip

\noindent {\bf Evaluation Protocol}: For the results in Figure~\ref{fig:bc-results} and Table~\ref{tab:all-stats}, the performance for each model and independent seed is quantified by computing a smoothed average of the average metrics over a fixed window during training (average metrics were evaluated very 1k training steps over 15 rollouts). The same window is used for all models (thus, all models have seen the same amount of training data) and we present the mean and $\pm 1 \sigma$ range for the computed smoothed values across three independent seeds. For the results in Figure~\ref{fig:bc-drop}, each model and seed, fine-tuned at $p_d=0.1$, is evaluated at varying drop probabilities within $[0, 0.5]$, where for each evaluation $p_d$, we recorded the average performance metrics $r_{\max}$ and $r_{\Sigma}$ over 100 rollouts. We then present the mean and $\pm 1\sigma$ range of these average metrics over 3 independent seeds.
\medskip

\noindent {\bf Computing Infrastructure}: All training was performed on Google v2 $2\times 2$ TPUs, and all code was written in JAX~\cite{jax2018github}, using the Flax neural network library~\cite{flax2020github}.

\subsection{Additional results}
\label{app:results}

Table~\ref{tab:all-stats} provides the raw performance numbers illustrated in Figure~\ref{fig:bc-results}, while Figure~\ref{fig:all-loss} illustrates the test loss as a function of training steps.

\begin{table*}
\subfloat[]{
\begin{tabular}{l ccc}
\toprule
\multirow{2}{*}{} & \multicolumn{3}{c}{CC} \\
\addlinespace[6pt]
\cmidrule(lr){2-4}\\
{\it Method} & 10 & 20 & 30 \\
\midrule
$\cde$ & $0.989 \pm 0.005$ & $0.987 \pm 0.002$ & $0.984 \pm 0.002$    \\
$\ndp$ & $0.989 \pm 0.004$ & $0.984 \pm 0.008$ & $0.923 \pm 0.017$      \\
$\stale$ & $0.984 \pm 0.003$ & $0.99 \pm 0.002$ & $0.99 \pm 0.001$       \\
$\mastale$ & $0.99 \pm 0.004$ & $0.99 \pm 0.002$  & $0.987 \pm 0.001$           \\
\bottomrule
\end{tabular}
}%
\hfill
\subfloat[]{
\begin{tabular}{l ccc}
\toprule
\multirow{2}{*}{} & \multicolumn{3}{c}{CC}\\
\addlinespace[6pt]
\cmidrule(lr){2-4}\\
{\it Method} & 10 & 20 & 30 \\
\midrule
$\cde$ & $37.06 \pm 0.86$ & $37.16 \pm 0.23$ & $37.35 \pm 0.67$      \\
$\ndp$ & $36.85 \pm 0.58$ & $36.46 \pm 0.58$  & $33.83 \pm 0.86$  \\
$\stale$ & $36.34 \pm 0.30$ & $37.25 \pm 0.16$ & $37.28 \pm 0.03$  \\
$\mastale$ & $36.97 \pm 0.36$& $37.13 \pm 0.34$  & $37.16 \pm 0.11$  \\
\bottomrule
\end{tabular}
}\\
\subfloat[]{
\resizebox{0.5\textwidth}{!}{%
\begin{tabular}{l cccc}
\toprule
\multirow{2}{*}{} & \multicolumn{4}{c}{BiC} \\
\addlinespace[6pt]
\cmidrule(lr){2-5}\\
{\it Method} & 2 & 4 & 6 & 8  \\
\midrule
$\cde$ & $0.975 \pm 0.02$ & $0.878 \pm 0.02$ & $0.745 \pm 0.007$ & $0.601 \pm 0.033$      \\
$\ndp$ & $0.976 \pm 0.004$ & $0.748 \pm 0.01$ &  $0.533 \pm 0.027$ & $0.378 \pm 0.052$    \\
$\stale$ & $0.948 \pm 0.006$ & $0.769 \pm 0.01$ & $0.56 \pm 0.085$  & $0.495 \pm 0.018$    \\
$\mastale$ & $0.943 \pm 0.002$ & $0.714 \pm 0.03$ & $0.6 \pm 0.019$ & $0.516 \pm 0.023$ \\
\bottomrule
\end{tabular}%
}
}%
\hfill
\subfloat[]{
\resizebox{0.5\textwidth}{!}{%
\begin{tabular}{l cccc}
\toprule
\multirow{2}{*}{} & \multicolumn{4}{c}{BiC} \\
\addlinespace[2pt]
\cmidrule(lr){2-5}\\
{\it Method} & 2 & 4 & 6 & 8  \\
\midrule
$\cde$ & $3.30 \pm 0.25$ & $2.35 \pm 0.19$ & $1.14 \pm 0.035$ & $-0.02 \pm 0.28$\\
$\ndp$ & $3.30 \pm 0.05$ & $1.21 \pm 0.11$ & $-0.4 \pm 0.2$  & $-1.65 \pm 0.395$   \\
$\stale$ & $3.09 \pm 0.07$ & $1.46 \pm 0.11$ & $-0.27 \pm 0.71$ & $-0.77 \pm 0.13$ \\
$\mastale$ & $2.81 \pm 0.07$ & $1.00 \pm 0.24$ & $0. \pm 0.15$ & $-0.63 \pm 0.232$    \\
\bottomrule
\end{tabular}%
}
} \\
\subfloat[]{
\resizebox{0.5\textwidth}{!}{%
\begin{tabular}{l cccc}
\toprule
\multirow{2}{*}{} & \multicolumn{4}{c}{vBiC} \\
\addlinespace[6pt]
\cmidrule(lr){2-5}\\
{\it Method} & 2 & 4 & 6 & 8  \\
\midrule
$\cde$ & $0.973 \pm 0.008$ & $0.861 \pm 0.002$ & $0.717 \pm 0.012$ & $0.617 \pm 0.007$ \\
$\ndp$ & $0.973 \pm 0.005$ & $0.817 \pm 0.01$ & $0.585 \pm 0.004$ & $0.394 \pm 0.019$  \\
$\stale$ & $0.86 \pm 0.013$ & $0.410 \pm 0.011$ & $0.349 \pm 0.002$  & $0.313 \pm 0.013$    \\
$\mastale$ & $0.920 \pm 0.008$ & $0.566 \pm 0.009$ & $0.411 \pm 0.007$ & $0.357 \pm 0.009$   \\
\bottomrule
\end{tabular}%
}
}%
\hfill
\subfloat[]{
\resizebox{0.5\textwidth}{!}{%
\begin{tabular}{l cccc}
\toprule
\multirow{2}{*}{} & \multicolumn{4}{c}{vBiC} \\
\addlinespace[3pt]
\cmidrule(lr){2-5}\\
{\it Method} & 2 & 4 & 6 & 8  \\
\midrule
$\cde$ & $3.30 \pm 0.1$ & $1.85 \pm 0.03$ & $0.452 \pm 0.1$ & $-0.55 \pm 0.05$  \\
$\ndp$ & $3.16 \pm 0.06$ & $1.20 \pm 0.02$ & $-0.856 \pm 0.02$ & $-2.35 \pm 0.15$    \\
$\stale$ & $1.69 \pm 0.15$ & $-2.16 \pm 0.08$ & $-2.65 \pm 0.04$ & $-2.98 \pm 0.11$  \\
$\mastale$ & $2.51 \pm 0.09$ & $-0.90 \pm 0.07$ & $-2.06 \pm 0.08$ & $-2.65 \pm 0.07$     \\
\bottomrule
\end{tabular}%
}
}
\caption{\footnotesize{Average (over 15 rollouts) $\rmax$ (left), and $\rtot$ (right), for varying values of $T$ across all environments. $\rmax=1$ indicates task success while larger values of $\rtot$ indicate a faster time to completion. The presented means and $\pm 1\sigma$ ranges are computed across 3 independent seeds.}}
\label{tab:all-stats}
\end{table*}

\begin{figure*}[h]
\centering
\begin{subfigure}[b]{0.32\textwidth}
 \centering
 \includegraphics[width=\textwidth]{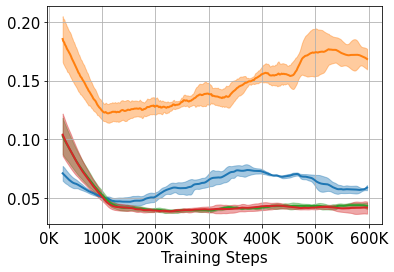}
\end{subfigure}
\begin{subfigure}[b]{0.32\textwidth}
 \centering
 \includegraphics[width=\textwidth]{cloth_loss_20.png}
\end{subfigure}
\begin{subfigure}[b]{0.32\textwidth}
 \centering
 \includegraphics[width=\textwidth]{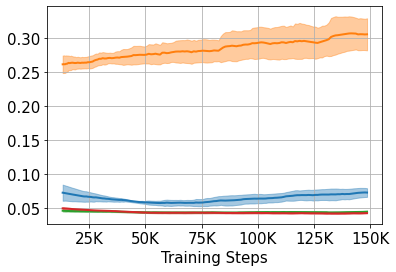}
\end{subfigure}\\
\begin{subfigure}[b]{0.24\textwidth}
 \centering
 \includegraphics[width=\textwidth]{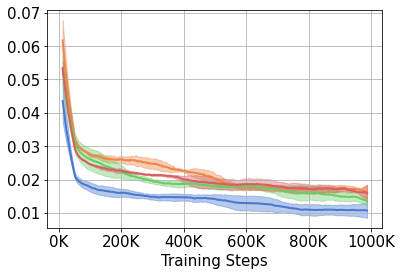}
\end{subfigure}
\begin{subfigure}[b]{0.24\textwidth}
 \centering
 \includegraphics[width=\textwidth]{bic_loss_4.png}
\end{subfigure}
\begin{subfigure}[b]{0.24\textwidth}
 \centering
 \includegraphics[width=\textwidth]{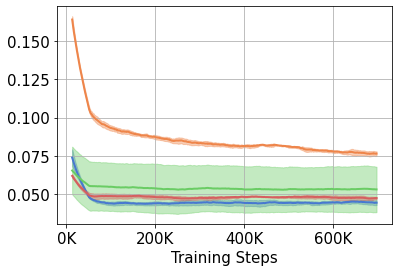}
\end{subfigure}
\begin{subfigure}[b]{0.24\textwidth}
 \centering
 \includegraphics[width=\textwidth]{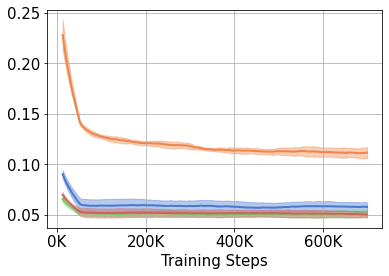}
\end{subfigure}\\
\begin{subfigure}[b]{0.24\textwidth}
 \centering
 \includegraphics[width=\textwidth]{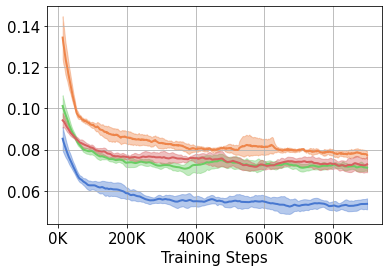}
\end{subfigure}
\begin{subfigure}[b]{0.24\textwidth}
 \centering
 \includegraphics[width=\textwidth]{vbic_loss_4.png}
\end{subfigure}
\begin{subfigure}[b]{0.24\textwidth}
 \centering
 \includegraphics[width=\textwidth]{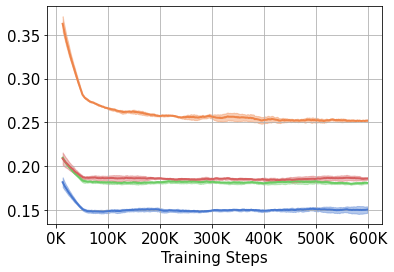}
\end{subfigure}
\begin{subfigure}[b]{0.24\textwidth}
 \centering
 \includegraphics[width=\textwidth]{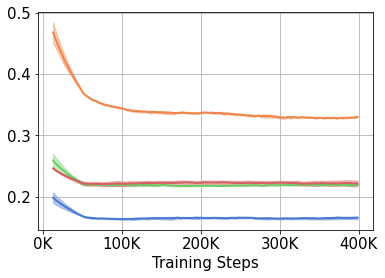}
\end{subfigure}
\caption{\footnotesize{Throttled-BC test loss vs training steps for cloth-covering (\emph{top}), ball-in-cup (\emph{middle}), and variable ball-in-cup (\emph{bottom}). $T$ is increasing from left-to-right.}}
\label{fig:all-loss}
\end{figure*}

}{}

\end{document}